\documentclass[twocolumn, 10pt]{article}
\usepackage[utf8]{inputenc}

\usepackage{stywhispers,amsmath,epsfig}
\usepackage{float}
\usepackage{times} 
\usepackage[colorlinks=true, linkcolor=blue, citecolor=blue, urlcolor=blue]{hyperref}
\usepackage{etoolbox}
\apptocmd{\thebibliography}{\setlength{\itemsep}{0pt}\setlength{\parskip}{0pt}}{}{}
\usepackage{titlesec}
\titlespacing*{\subsubsection}{0pt}{0.5\baselineskip}{0.3\baselineskip}

\title{Benchmarking Foundation Models for Hyperspectral Image Classification: Application to Cereal Crop Type Mapping}

\name{Walid Elbarz$^{1}$\thanks{Corresponding author: walid.ELBARZ@um6p.ma}, Mohamed Bourriz$^{1,2 ,3}$, Hicham Hajji$^4$, Hamd Ait Abdelali$^1$, François Bourzeix$^1$}
\address{
$^1$Analytics Lab (A-Lab), UM6P, Rabat 11103, Morocco\\
$^2$Center for Remote Sensing Applications (CRSA), UM6P, Ben Guerir 43150, Morocco\\
$^3$Friedrich Schiller University Jena, Department of Geography, Jena 07743, Germany \\
$^4$Dept. of Cartography and Photogrammetry, IAV Hassan II, Rabat 10101, Morocco
}
\begin{document}
%
\maketitle
\begin{abstract}
This study benchmarks three foundation models for cereal crop mapping using hyperspectral imagery: \textbf{HyperSigma}, \textbf{DOFA}, and Vision Transformers pre-trained on the \textbf{SpectralEarth} dataset---a large multitemporal hyperspectral archive. The models were fine-tuned using manually labeled data from a training region, while evaluation was performed on an independent test region . Performance was assessed using overall accuracy (OA), average accuracy (AA), and F1-score metrics. \textbf{HyperSigma} achieved an OA of \textbf{34.5\%} ($\pm$1.8\%), \textbf{DOFA} reached \textbf{62.6\%} ($\pm$3.5\%), and the \textbf{SpectralEarth} model outperformed both with an OA of \textbf{93.5\%} ($\pm$0.8\%). Notably, a compact SpectralEarth variant trained from scratch achieved \(91\%\), highlighting the importance of model architecture in enabling strong generalization across different geographic regions and sensor platforms.

\end{abstract}
\begin{keywords}
Remote sensing, Cross-region evaluation, Foundation models, Hyperspectral imaging, Crop mapping
\end{keywords}
\section{Introduction}
\label{sec:intro}

Hyperspectral imagery has emerged as a promising remote sensing modality due to its rich spectral information spanning hundreds of bands and offering much higher spectral resolution than multispectral images. This richness has enabled advances in diverse domains—particularly in precision agriculture—\cite{Bhargava2024,thenkabail2013hyperspectral,hughes1968mean}. However, its high dimensionality increases computational intensity and often leads to the curse of dimensionality—commonly termed the Hughes phenomenon. These issues are  addressed through dimensionality reduction techniques such as Principal Component Analysis (PCA), as used in \cite{wang_hypersigma_2024} for model pretraining, or by selecting Optimal Hyperspectral Narrow Bands (OHNB) \cite{Aneece2022}.

Deep learning methods have shown potential in tackling these challenges; however, they require large quantities of costly annotated data. Recently, geospatial foundation models—large-scale architectures pretrained using self-supervised learning on vast remote sensing corpora—have shown promise in overcoming these limitations \cite{prithvi_2.0}. Yet, most model evaluations have been conducted on idealized benchmarks that do not reflect real-world conditions. For instance, strong model performance has been demonstrated on hyperspectral image classification benchmark datasets \cite{wang_hypersigma_2024,Roy2019-tt,Feng2023}. However, these benchmarks are prone to data leakage if improperly configured.

Comparative studies on hyperspectral-specific foundation models remain scarce, particularly within the context of crop mapping. The application of such models in this domain has only gained attention recently—most notably in the review by \cite{Bourriz2025_ai}, which outlined emerging AI trends in hyperspectral crop mapping and emphasized the potential of foundation models as a novel methodology. A very recent study \cite{nedungadi2025foundational} applied foundation models in practice but notably excluded hyperspectral imagery as a modality. Moreover, no prior work has directly assessed these models under real-world crop mapping scenarios based on hyperspectral images. This study seeks to fill these gaps through the following contributions:
\begin{itemize}\setlength\itemsep{0em}
    \item Adaptation of foundation models to pixel-level classification using a simple upsampling layer.
    \item A benchmark of hyperspectral foundation models using real-world, limited annotated data across distinct regions and hyperspectral satellite sensors.
\end{itemize}
\begin{figure*}[t]
  \centering
  \includegraphics[width=0.99\textwidth]{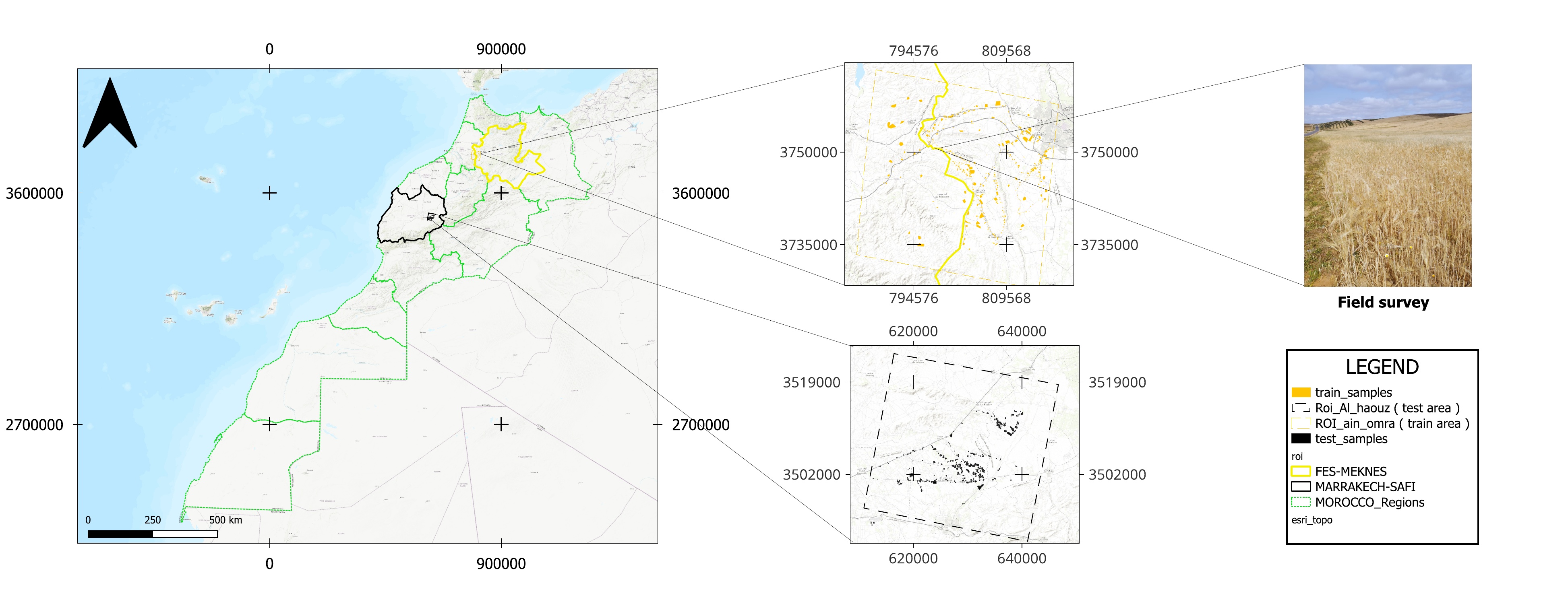} 
  \caption{Study area with the training and testing ROIs.}
  \label{fig:study_area}
\end{figure*}

\section{Methodology}
\subsection{Study area}

\label{ssec:subhead}
This study takes place in Morocco. As Figure~\ref{fig:study_area} demonstrates, two regions of interest (ROIs) were selected. The first ROI originates from the \textbf{Aïn Orma} commune, located in the \textbf{Fès-Meknès} region, whereas the second ROI is in the \textbf{Al Haouz} plains, located in the \textbf{Marrakech-Safi} region. The two ROIs differ in terms of climate as well as topography, making them exemplary candidates to test scalability. For simplicity and readability, we denote \textbf{Aïn Orma} as the training zone and \textbf{Al Haouz} as the test zone.

\subsection{Hyperspectral  data }
Hyperspectral images (HSIs) serve as the input data for the foundation models in this study. For the \textbf{training zone}, we utilized hyperspectral data acquired by the \textit{Environmental Mapping and Analysis Program (EnMAP)} satellite on \textbf{2025-03-21}, while the \textbf{testing zone} relied on data captured by \textit{PRISMA (PRecursore IperSpettrale della Missione Applicativa)} on \textbf{2025-03-30}. Since these two sensors differ in their spectral ranges and sampling characteristics, the PRISMA data was spectrally resampled to match the EnMAP wavelength configuration. \cite{bourriz_intercomparison_2025} demonstrated a high correlation between EnMAP and PRISMA data, suggesting that, in theory, models should generalize across the two different sensors. 
During preprocessing, bands containing NaN values were removed from both HSIs, resulting in 167 valid spectral bands. Both training and testing images were smoothed using Savitzky–Golay filter . Subsequently, the images were co-registered to a reference Sentinel-2 image to ensure accurate georeferencing aligned with the ground truth mask.
\subsection{Ground truth data }
Ground truth data was collected manually through field surveys using the QField software. Within each ROI, samples were identified,digitized, and labeled according to the crop type. The labels were harmonized into two broad categories: \textbf{class 1} for cereal, and \textbf{class 2} for non-cereal. After cleaning the data in terms of geometric consistency and attribute correctness, the labeled polygons were rasterized to generate ground truth masks. These masks were then used for both training and testing the selected models.


\label{sec:format}

\subsection{Foundation models}
\label{sec:pagestyle}
Foundation models are large-scale architectures pretrained on vast datasets, enabling broad generalization across downstream tasks with minimal architectural adjustments \cite{wang_hypersigma_2024}. In computer vision, they are typically trained using self-supervised methods such as the Masked Autoencoder (MAE) paradigm \cite{he_masked_2021}, where parts of the input image are masked and reconstructed. This approach has been successfully adapted to remote sensing, with spectral and even temporal masking applied in multispectral and hyperspectral contexts \cite{tseng_lightweight_2024}. Below, we  describe the foundation models used in this benchmark and the fine-tuning methodology .
\subsubsection{Models selection}
The benchmark targets foundation models pretrained on hyperspectral imagery or including it as a training modality. Three criteria guided the selection: (1) models must use hyperspectral data during pretraining, excluding multispectral-only models such as Prithvi~2.0 \cite{prithvi_2.0} ; (2) pretrained weights must be publicly available to ensure reproducibility; and (3) models should offer flexibility for diverse downstream tasks without major architectural changes. Based on these criteria, three representative models were selected:

\textbf{HyperSigma} (Hyperspectral Intelligence Comprehension Foundation Model) \cite{wang_hypersigma_2024} is a  model built by fusing two independently pretrained spectral and spatial subnetworks. It employs Sparse Sample Attention (SSA) to selectively focus on informative regions and is pretrained on HyperGlobal-450K, a dataset of 447k EO-1 and GF-5 image patches.

\textbf{DOFA (Dynamic One-For-All)} \cite{xiong_neural_2024} introduces a wavelength-conditioned hypernetwork and gated cross-modal attention to unify multiple sensing modalities. It is pretrained on a massive multimodal dataset comprising Sentinel-1 (4.6M), Sentinel-2 (978k), Gaofen, aerial RGB (2.3M USA samples), and EnMAP hyperspectral (11.5k) imagery.

\textbf{SpectralEarth} \cite{Braham2024-ov} is both a large-scale hyperspectral dataset and a set of pretrained foundation models derived from it. The dataset comprises over 538k image patches from 415k unique global locations, often with multi-temporal coverage, thereby surpassing the scale of HyperGlobal-450K. For benchmarking, we employ the Transformer-based foundation models released by the authors, which are specifically adapted for hyperspectral data through a spectral adapter.

\subsubsection{Fine-tuning Methodology}
The pixel-based classification approach assigns a unique label to each individual pixel. To incorporate spatial context, a small patch of size \(3 \times 3\) pixels centered on the target pixel is extracted, as supported by the ablation study \ref{sec:ablation_study_section}. Since the transformer backbone employs a patch size of \(4 \times 4\), direct patchification on a \(3 \times 3\) grid is not feasible. Therefore, the extracted patches are upsampled to \(16 \times 16\) pixels using bicubic interpolation, ensuring compatibility with the transformer’s patch embedding mechanism while keeping the input size computationally efficient for inference on full images. These upsampled patches serve as the model input.

\begin{figure}[h!]
  \centering
  \includegraphics[width=\linewidth]{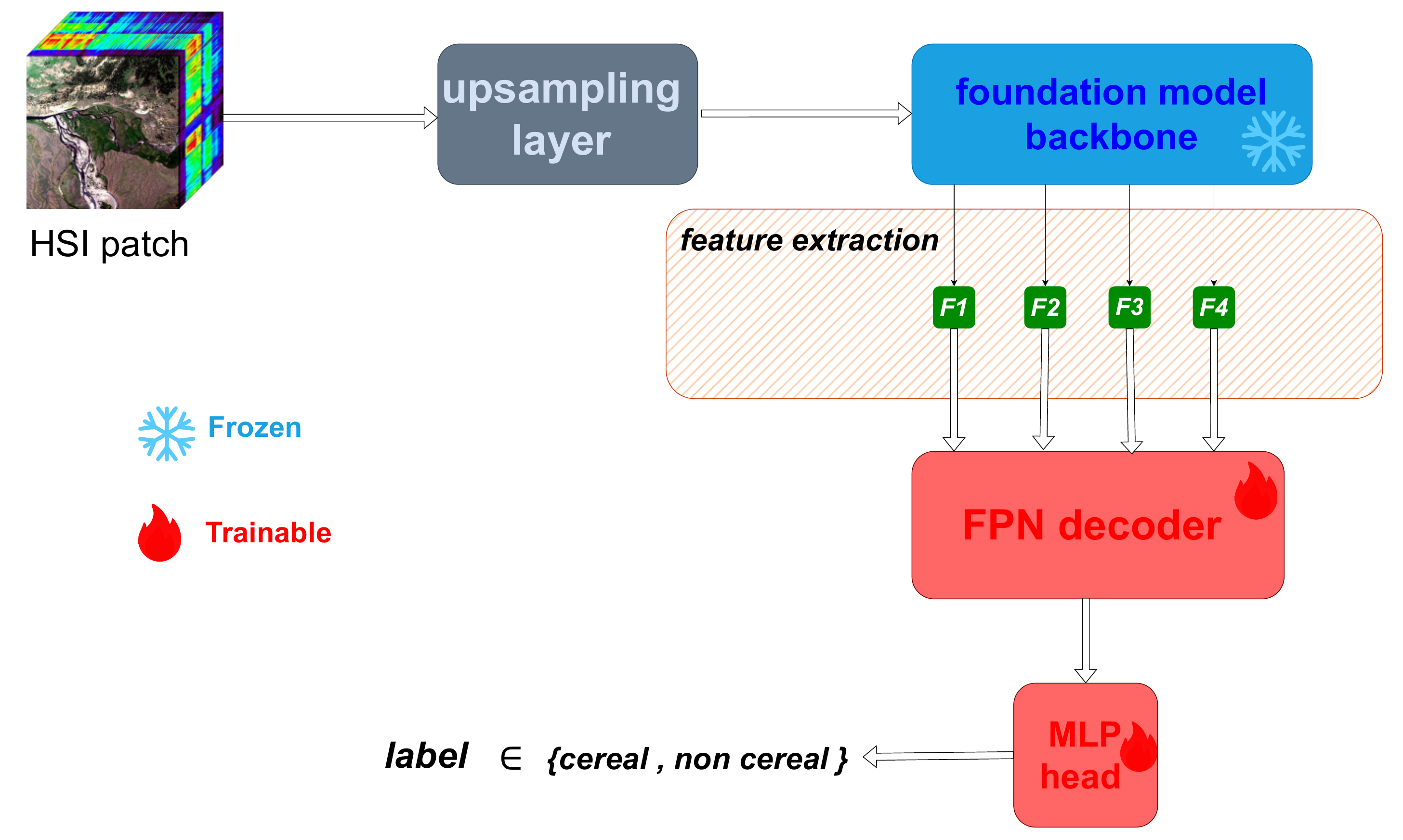} 
  \caption{Benchmark’s modeling pipeline.}
  \label{fig:modeling framework}
\end{figure}

For all experiments, the base versions of the selected models are used. A consistent decoder architecture, used for pixel-based classification, adapted from \cite{wang_hypersigma_2024}, is employed across all models to ensure a fair comparison. The resulting architecture is summarized in figure \ref{fig:modeling framework}, we keep the backbone's weights frozen during training. The number of parameters of the resulting classification model is summarized in table \ref{tab:foundation_models}.


Model training used the \textbf{AdamW optimizer} with weight decay and a \textbf{cosine annealing schedule} for stable convergence, while \textbf{cross-entropy loss} served as the objective to minimize classification errors.

\begin{table}[t]
\centering
\caption{Summary of selected foundation models.}
\label{tab:foundation_models}
\begin{tabular}{lcc}
\hline
\textbf{Model Name} & \textbf{Model Size} & \textbf{Parameter Count (M)} \\
\hline
DOFA               & Base                & 89 \\
HyperSigma           & Base                & 183 \\
SpectralEarth-ViT    & Base                & 88 \\
\hline
\end{tabular}
\end{table}
\label{sec:typestyle}
\section{Results}
The \textbf{SpectralEarth} model demonstrated the strongest performance across all evaluation metrics, with an overall accuracy (OA) of \textbf{93.5\%} ($\pm$0.8\%), average accuracy (AA) of \textbf{93.4\%} ($\pm$0.8\%), a kappa coefficient of \textbf{0.85} ($\pm$0.01), and an F1-score for the cereal class of \textbf{90\%} ($\pm$0.9\%). These results indicate consistent classification performance across crop types and a high ability to accurately identify cereals. \textbf{DOFA} produced moderate results, with an OA of \textbf{62.6\%} ($\pm$3.5\%), AA of \textbf{72.6\%} ($\pm$2.4\%), and a kappa coefficient of \textbf{0.34} ($\pm$0.04). While it achieved a high recall of 98\% for cereals in the original text, the new F1-score of \textbf{62.4\%} ($\pm$2.0\%) suggests a moderate balance between precision and recall. \textbf{HyperSigma} showed the weakest performance, with an OA of \textbf{34.5\%} ($\pm$1.8\%), AA of \textbf{52.4\%} ($\pm$1.3\%), and a kappa value of \textbf{0.03} ($\pm$0.02), indicating minimal agreement with ground truth. The F1-score for cereals was \textbf{48.8\%} ($\pm$0.7\%) these metrics reflect poor alignment with the ground truth and limited capacity to distinguish between the two classes, indicating that the model fails to generalize effectively in this classification setting.

\begin{figure}[h!]
    \centering
    \includegraphics[width=\linewidth]{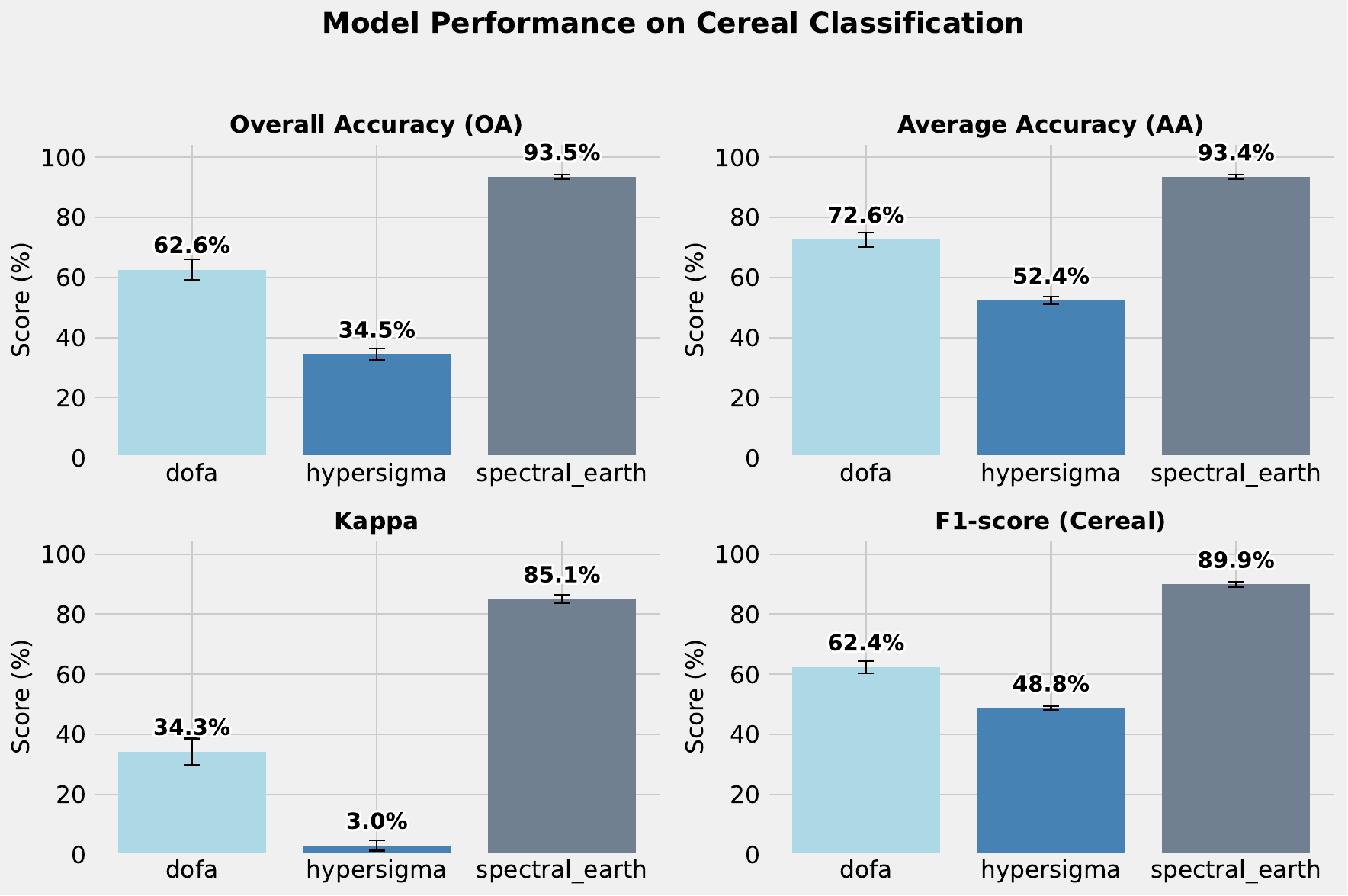}
    \caption{Comparison of SpectralEarth, DOFA, and HyperSigma on the local data  using OA, AA, Kappa and F1 score (cereal) metrics.}
    \label{fig:morocco_model_performance}
\end{figure}
\begin{figure}[h!]
    \centering
    \begin{minipage}[b]{0.45\linewidth}
        \centering
        \includegraphics[width=\linewidth, height=3cm]{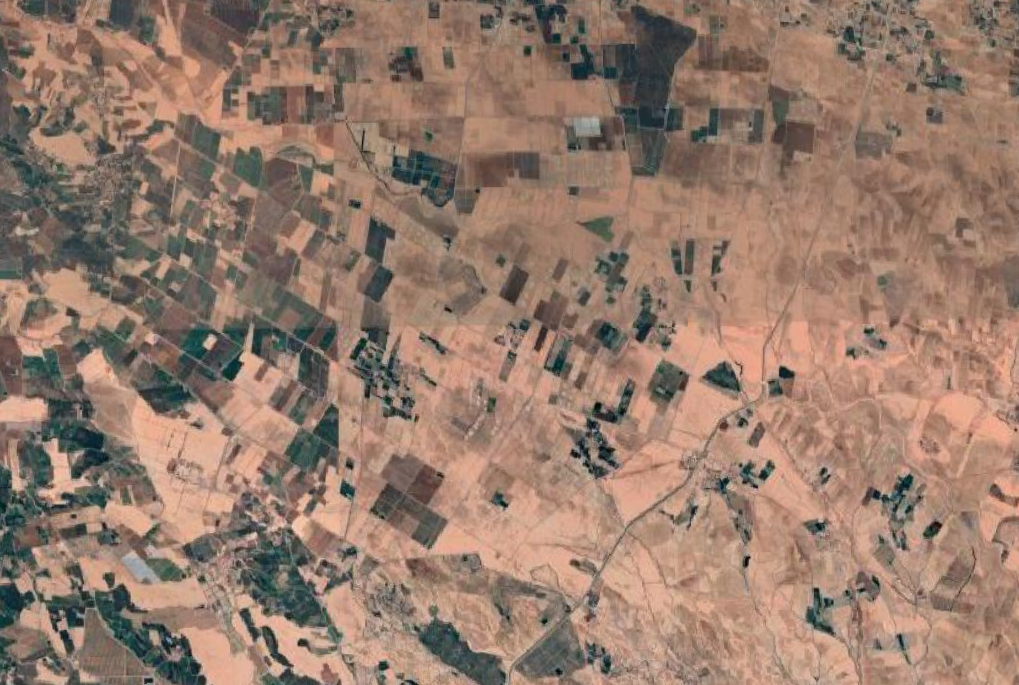}
        \centerline{(a) Reference image}\medskip
    \end{minipage}
    \hfill
    \begin{minipage}[b]{0.45\linewidth}
        \centering
        \includegraphics[width=\linewidth, height=3cm]{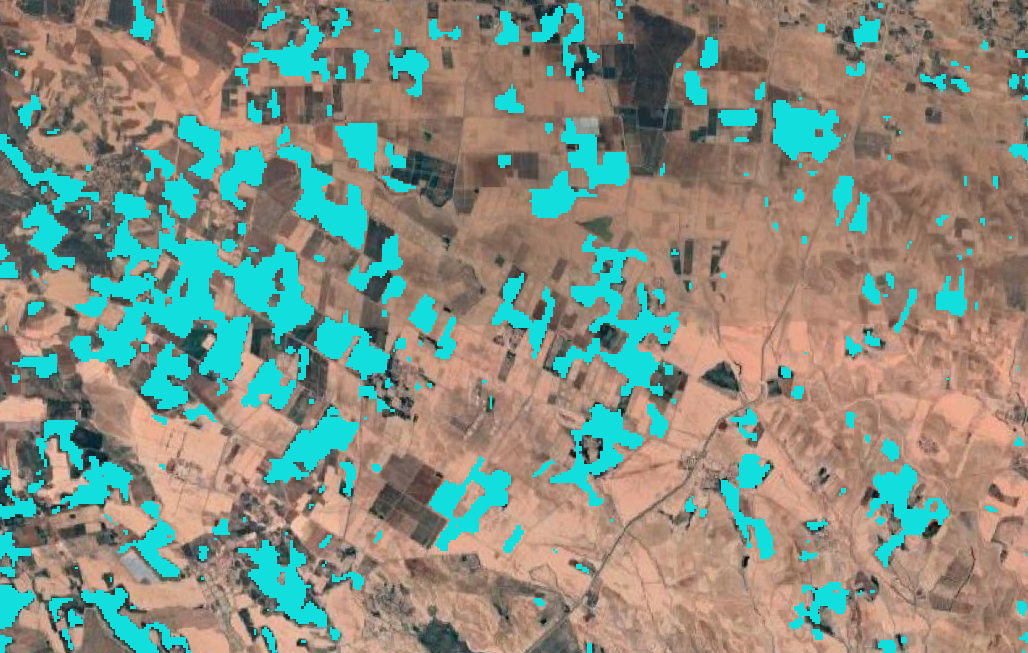}
        \centerline{(b) SpectralEarth}\medskip
    \end{minipage}

    \vspace{0.5cm}

    \begin{minipage}[b]{0.45\linewidth}
        \centering
        \includegraphics[width=\linewidth, height=3cm]{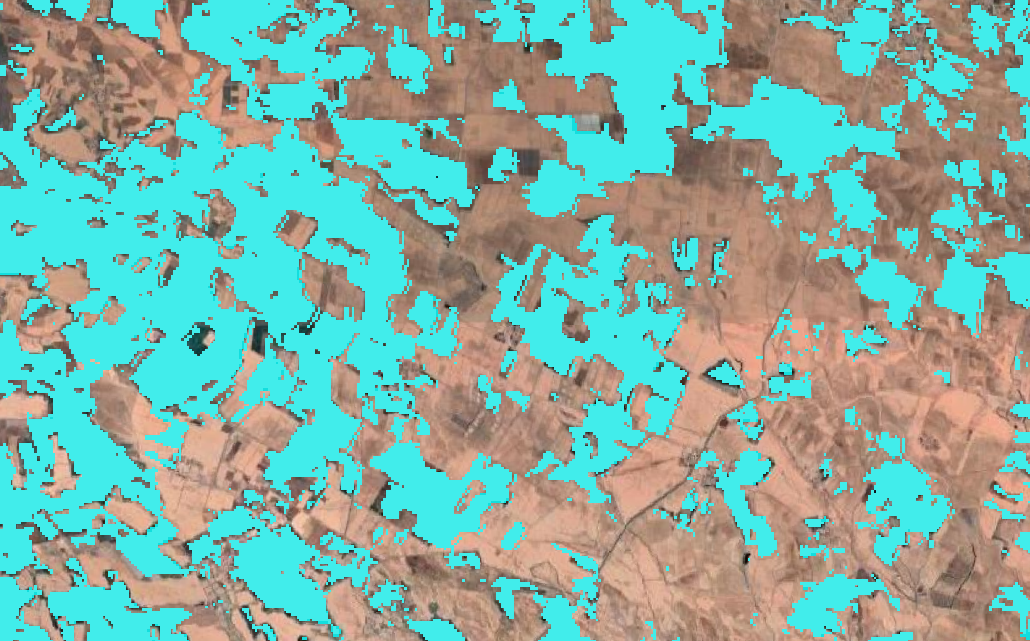}
        \centerline{(c) DOFA}\medskip
    \end{minipage}
    \hfill
    \begin{minipage}[b]{0.45\linewidth}
        \centering
        \includegraphics[width=\linewidth, height=3cm]{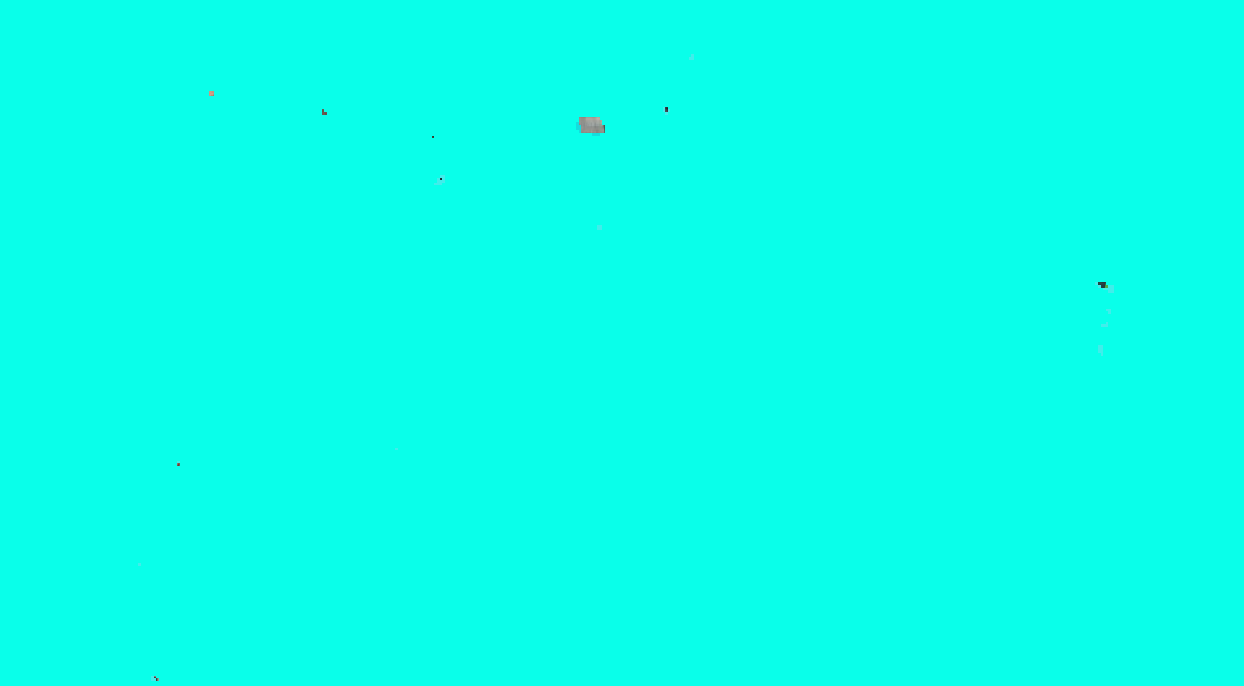}
        \centerline{(d) HyperSigma}\medskip
    \end{minipage}

    \caption{Visual comparison of model predictions on a subset of the test area. The reference image (a) is shown alongside predictions from SpectralEarth (b), DOFA (c), and HyperSigma (d). the cyan color denotes  class 1 (cereal) predicted by the model  }
    \label{fig:morocco_visual_comparison}
\end{figure}

The results are further illustrated in Figure~\ref{fig:morocco_visual_comparison}, where the cereal masks generated by each model are overlaid on a satellite basemap. These results do not imply that SpectralEarth is the best model overall; rather, they demonstrate its adaptability to this specific scenario, characterized by scarce and limited labels.
\section{Ablation study}
\label{sec:ablation_study_section}
This section evaluates the impact of varying patch sizes centered on each pixel before the upsampling layer, as well as the model’s performance without upsampling. The SpectralEarth-ViT model was trained from scratch on the dataset with reduced parameters, as detailed in Table~\ref{tab:spectralearth_comparison}, and is hereafter referred to as the nano-sized model. Patch sizes of 1, 3, 5, 11, and 28 pixels were tested. For the largest patch size 28, upsampling was omitted because its dimensions exceeded the target size for upsampling.

\begin{figure}[h!]
    \centering
    \includegraphics[width=\linewidth]{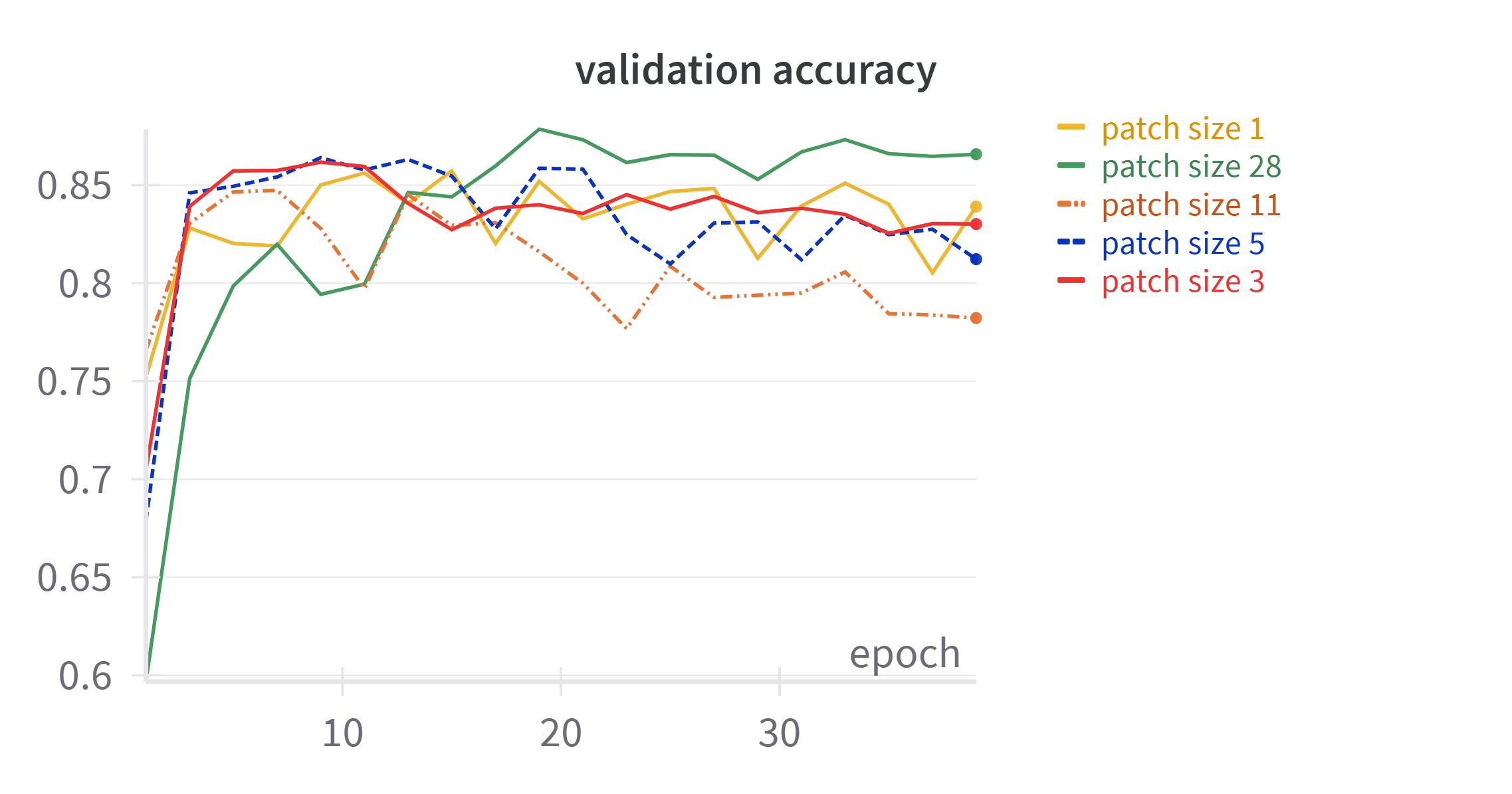}
    \caption{SpectralEarth Nano ViT’s validation accuracy using different patch sizes.}
    \label{fig:validation accuracy of SpectralEarth Nano models.}
\end{figure}

\begin{figure}[h]
    \centering
    \includegraphics[width=\linewidth]{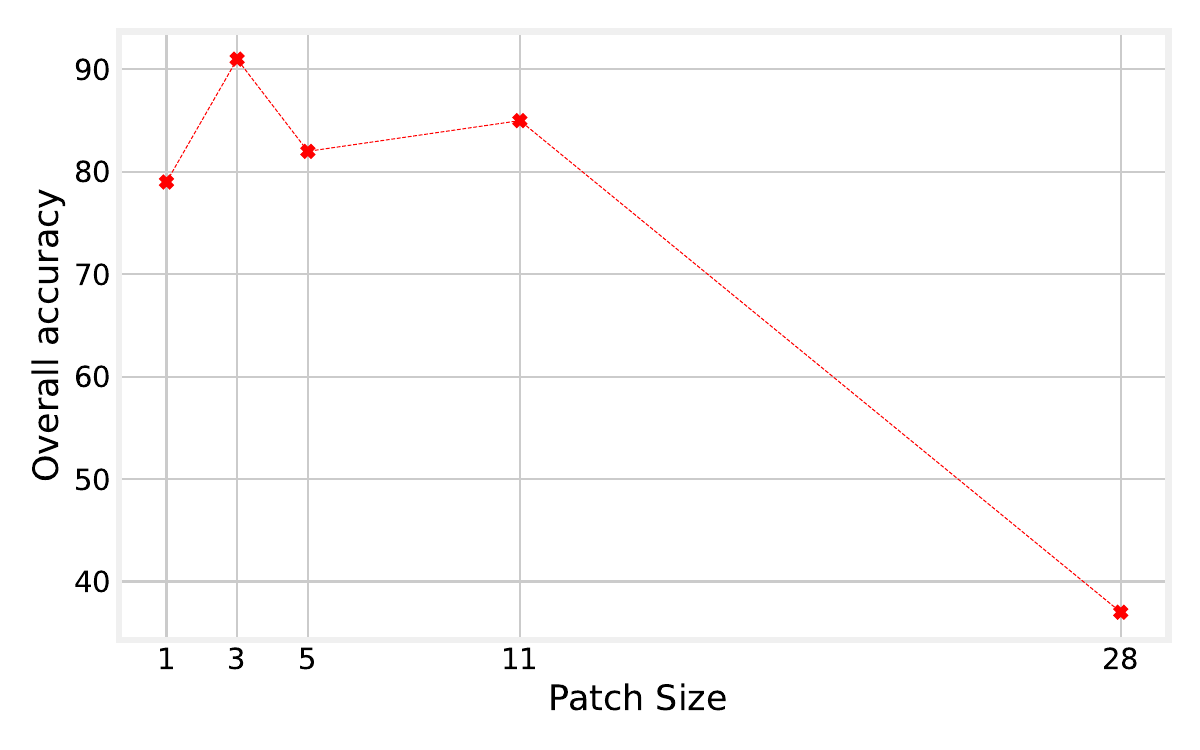}
    \caption{Plot of overall accuracy of spectral earth model on the test set  using different patch sizes, the patch size of 28 performed worst, whereas the patch size of 3 achieved the best performance while using only 3\% of the parameters of the pretrained base model.}
    \label{fig:ablation study}
\end{figure}

Validation accuracy results, shown in Figure~\ref{fig:validation accuracy of SpectralEarth Nano models.}, indicate that patch sizes 1, 3, and 5 achieved similar performance around 86\%, while patch size 11 reached 85\%. The largest patch size 28 attained the highest validation accuracy of 87\%. However, test accuracy revealed a different pattern, as shown in Figure~\ref{fig:ablation study}: patch size 3 yielded the best results, whereas patch size 28 dropped sharply to 37\%. This suggests that larger patch sizes do not necessarily improve generalization, despite their high validation accuracy.

\begin{table}[H]
\centering
\caption{Comparison of Nano and Base SpectralEarth-ViT models in model size and training time per epoch on a single NVIDIA L40S GPU}
\label{tab:spectralearth_comparison}
\resizebox{\columnwidth}{!}{%
\begin{tabular}{lcc}
\hline
\textbf{Model} & \textbf{Parameters (M)} & \textbf{Training Time / Epoch (seconds)} \\
\hline
Vit-Nano  & 1.4 & 4 \\
Vit-Base & 87.9 & 32 \\
\hline
\end{tabular}
}
\end{table}

The poor performance of the patch size 28 can be attributed to two main factors. First, data leakage occurred because some validation pixels overlapped with training patches at this large scale. In contrast, smaller patches (e.g., size 3) minimized such overlap, reducing leakage. Second, from an architectural perspective, the transformer model’s wide receptive field captures extensive spatial context. Providing excessive spatial content (as with large patches) introduces background noise, hindering scalability. Conversely, using very small patches (e.g., size 1) limits spatial information, reducing model performance. The patch size 3 represents a balanced choice, offering sufficient spatial detail without excessive noise.

\section{Conclusion }
Foundation models hold strong potential for crop mapping with hyperspectral data, yet key challenges remain—especially in data-scarce regions and data-agnostic settings. Their success depends on protocols that avoid data leakage, use generalizable architectures, and ensure reproducible evaluation. This work highlights the need for scalable assessment across regions and sensors. Future research should explore multimodal fusion (e.g., multispectral, SAR, thermal) to enhance crop-type discrimination, and develop methods that suppress background noise while preserving spatial detail for more accurate classification.
\section{Acknowledgments}
This work was conducted within the Yield Gap project (OCP Foundation–UM6P agreement) and made use of PRISMA and EnMAP data provided by the Italian Space Agency (ASI) and the German Aerospace Center (DLR).


\bibliographystyle{IEEEbib}
\bibliography{ref3} 

\begin{thebibliography}{10}

\bibitem{Bhargava2024}
Anuja Bhargava, Ashish Sachdeva, Kulbhushan Sharma, Mohammed~H. Alsharif, Peerapong Uthansakul, and Monthippa Uthansakul,
\newblock ``Hyperspectral imaging and its applications: A review,''
\newblock {\em Heliyon}, vol. 10, no. 12, pp. e33208, 2024.

\bibitem{thenkabail2013hyperspectral}
Prasad~S Thenkabail,
\newblock ``Hyperspectral remote sensing of vegetation and agricultural crops,''
\newblock {\em Photogrammetric Engineering \& Remote Sensing}, vol. 79, no. 7, pp. 697--709, 2013.

\bibitem{hughes1968mean}
G.~F. Hughes,
\newblock ``On the mean accuracy of statistical pattern recognizers,''
\newblock {\em IEEE Transactions on Information Theory}, vol. 14, no. 1, pp. 55--63, 1968.

\bibitem{wang_hypersigma_2024}
Di~Wang, Meiqi Hu, Yao Jin, Yuchun Miao, Jiaqi Yang, Yichu Xu, Xiaolei Qin, Jiaqi Ma, Lingyu Sun, Chenxing Li, Chuan Fu, Hongruixuan Chen, Chengxi Han, Naoto Yokoya, Jing Zhang, Minqiang Xu, Lin Liu, Lefei Zhang, Chen Wu, Bo~Du, Dacheng Tao, and Liangpei Zhang,
\newblock ``{HyperSIGMA}: {Hyperspectral} {Intelligence} {Comprehension} {Foundation} {Model},'' June 2024,
\newblock arXiv:2406.11519 [cs].

\bibitem{Aneece2022}
Itiya Aneece and Prasad~S. Thenkabail,
\newblock ``New generation hyperspectral sensors desis and prisma provide improved agricultural crop classifications,''
\newblock {\em Photogrammetric Engineering and; Remote Sensing}, vol. 88, no. 11, pp. 715–729, Nov. 2022.

\bibitem{prithvi_2.0}
Daniela Szwarcman, Sujit Roy, Paolo Fraccaro, Thorsteinn~Eli Gislason, Benedikt Blumenstiel, Rinki Ghosal, Pedro~Henrique de~Oliveira, Joao Lucas de~Sousa Almeida, Rocco Sedona, Yanghui Kang, Srija Chakraborty, Sizhe Wang, Carlos Gomes, Ankur Kumar, Myscon Truong, Denys Godwin, Hyunho Lee, Chia-Yu Hsu, Ata~Akbari Asanjan, Besart Mujeci, Disha Shidham, Trevor Keenan, Paulo Arevalo, Wenwen Li, Hamed Alemohammad, Pontus Olofsson, Christopher Hain, Robert Kennedy, Bianca Zadrozny, David Bell, Gabriele Cavallaro, Campbell Watson, Manil Maskey, Rahul Ramachandran, and Juan~Bernabe Moreno,
\newblock ``Prithvi-eo-2.0: A versatile multi-temporal foundation model for earth observation applications,'' 2024.

\bibitem{Roy2019-tt}
Swalpa~Kumar Roy, Gopal Krishna, Shiv~Ram Dubey, and Bidyut~B Chaudhuri,
\newblock ``{HybridSN}: Exploring {3D-2D} {CNN} feature hierarchy for hyperspectral image classification,''
\newblock {\em arXiv [cs.CV]}, 2019.

\bibitem{Feng2023}
Hao Feng, Yongcheng Wang, Zheng Li, Ning Zhang, Yuxi Zhang, and Yunxiao Gao,
\newblock ``Information leakage in deep learning-based hyperspectral image classification: A survey,''
\newblock {\em Remote Sensing}, vol. 15, no. 15, pp. 3793, July 2023.

\bibitem{Bourriz2025_ai}
Mohamed Bourriz, Hicham Hajji, Ahmed Laamrani, Nadir Elbouanani, Hamd~Ait Abdelali, Fran\c{c}ois Bourzeix, Ali El-Battay, Abdelhakim Amazirh, and Abdelghani Chehbouni,
\newblock ``Integration of hyperspectral imaging and ai techniques for crop type mapping: Present status, trends, and challenges,''
\newblock {\em Remote Sensing}, vol. 17, no. 9, pp. 1574, Apr. 2025.

\bibitem{nedungadi2025foundational}
Vishal Nedungadi, Xingguo Xiong, Aike Potze, Ron Van~Bree, Tao Lin, Marc Rußwurm, and Ioannis~N. Athanasiadis,
\newblock ``From general to specialized: The need for foundational models in agriculture,'' 2025.

\bibitem{bourriz_intercomparison_2025}
Mohamed Bourriz, Ahmed Laamrani, Ali El-Battay, Hicham Hajji, Nadir Elbouanani, Hamd~Ait Abdelali, François Bourzeix, Abdelhakim Amazirh, and Abdelghani Chehbouni,
\newblock ``An {Intercomparison} of {Two} {Satellite}-{Based} {Hyperspectral} {Imagery} ({PRISMA} and; {EnMAP}) for {Agricultural} {Mapping}: {Potential} of these sensors to produce hyperspectral time-series essential for tracking crop phenology and enhancing crop type mapping,''
\newblock Tech. {R}ep. EGU25-18417, Copernicus Meetings, Mar. 2025,
\newblock Conference Name: EGU25.

\bibitem{he_masked_2021}
Kaiming He, Xinlei Chen, Saining Xie, Yanghao Li, Piotr Dollár, and Ross Girshick,
\newblock ``Masked {Autoencoders} {Are} {Scalable} {Vision} {Learners},'' Dec. 2021,
\newblock arXiv:2111.06377 [cs].

\bibitem{tseng_lightweight_2024}
Gabriel Tseng, Ruben Cartuyvels, Ivan Zvonkov, Mirali Purohit, David Rolnick, and Hannah Kerner,
\newblock ``Lightweight, {Pre}-trained {Transformers} for {Remote} {Sensing} {Timeseries},'' Feb. 2024,
\newblock arXiv:2304.14065 [cs].

\bibitem{xiong_neural_2024}
Zhitong Xiong, Yi~Wang, Fahong Zhang, Adam~J. Stewart, Joëlle Hanna, Damian Borth, Ioannis Papoutsis, Bertrand~Le Saux, Gustau Camps-Valls, and Xiao~Xiang Zhu,
\newblock ``Neural {Plasticity}-{Inspired} {Multimodal} {Foundation} {Model} for {Earth} {Observation},'' June 2024,
\newblock arXiv:2403.15356 [cs].

\bibitem{Braham2024-ov}
Nassim Ait~Ali Braham, Conrad~M Albrecht, Julien Mairal, Jocelyn Chanussot, Yi~Wang, and Xiao~Xiang Zhu,
\newblock ``{SpectralEarth}: Training hyperspectral foundation models at scale,''
\newblock 2024.

\end{thebibliography}
\end{document}